\documentclass{article}

\usepackage[left=2.40cm, right=2.40cm, top=2.40cm, bottom=2.40cm]{geometry}

\usepackage[titletoc,title]{appendix}

\usepackage{amsfonts}       
\usepackage{graphicx}
\usepackage{amsmath}        
\usepackage{amsthm}
\usepackage{amssymb}
\usepackage{bm}             

\title{Sparse Coding by Spiking Neural Networks: \\
Convergence Theory and Computational Results}

\author{
	Ping Tak Peter Tang, Tsung-Han Lin, and Mike Davies \\ \\
    Intel Corporation \\
	\{peter.tang, tsung-han.lin, mike.davies\}@intel.com     
}

\date{}

\newcommand{\VS}{}

\newcommand{\bydef}{\stackrel{\mathrm{def}}{=}}

\newcommand{\tendsto}{\rightarrow}
\newcommand{\conv}{\ast}

\newtheorem{mythm}{Theorem}

\newtheorem{myprop}{Proposition}
\newtheorem*{myproof}{Proof}
\newtheorem{myassumption}{Assumption}

\newcommand{\reals}{\mathbb{R}}

\newcommand{\suchthat}{\,\mid\,}

\newcommand{\realsN}{\reals^N}
\newcommand{\realsM}{\reals^M}

\newcommand{\Enorm}[1]{\| {#1} \|_2}

\newcommand{\Onenorm}[1]{\| {#1} \|_1}
\newcommand{\inverse}[1]{{#1}^{-1}}
\newcommand{\transpose}[1]{{#1}^{T}}
\newcommand{\dist}[2]{\mathrm{dist}({#1},{#2})}
\newcommand{\ssup}[2]{{#1}^{(#2)}}

\newcommand{\vv}[1]{\mathbf{#1}}

\newcommand{\vva}{\vv{a}}
\newcommand{\vvb}{\vv{b}}

\newcommand{\vvr}{\vv{r}}
\newcommand{\vvs}{\vv{s}}
\newcommand{\vvu}{\vv{u}}

\newcommand{\vvx}{\vv{x}}
\newcommand{\vvy}{\vv{y}}

\newcommand{\vvzero}{\vv{0}}

\newcommand{\cc}[1]{{\cal #1}}
\newcommand{\ccA}{\cc{A}}

\newcommand{\ccC}{\cc{C}}

\newcommand{\ccI}{\cc{I}}

\newcommand{\ccM}{\cc{M}}

\newcommand{\ccS}{\cc{S}}

\newcommand{\vvastar}{\vva^*}

\newcommand{\vvustar}{\vvu^*}

\newcommand{\thres}{T}
\newcommand{\threslasso}{\thres_{\pm\lambda}}
\newcommand{\thresclasso}{\thres_\lambda}
\newcommand{\vvthres}{\vv{\thres}}

\newcommand{\vvphi}{{\boldsymbol\phi}}

\newcommand{\vvF}{\vv{F}}

\newcommand{\ustart}{\ssup{\vv{u}}{0}}

\newcommand{\fixedpts}{{\cal F}}

\newcommand{\fixedregion}{\hat{\fixedpts}}

\newcommand{\curr}{\mu}
\newcommand{\inputcurr}{b}
\newcommand{\currbias}{\lambda}
\newcommand{\poten}{v}
\newcommand{\potenthres}{\nu}
\newcommand{\potenfire}{\nu_f}
\newcommand{\potenreset}{\nu_r}
\newcommand{\spike}{\delta}
\newcommand{\spiketrain}{\sigma}
\newcommand{\avgcurr}{u}
\newcommand{\spikerate}{a}
\newcommand{\spiketime}[2]{t_{#1,#2}}
\newcommand{\starttime}{t_0}
\newcommand{\decay}{\alpha}
\newcommand{\expwidth}{\tau}
\newcommand{\expdecay}[1]{e^{-{#1}/\expwidth}}
\newcommand{\dirac}{\delta}
\newcommand{\spt}[2]{t_{#1,#2}}

\begin{document}

\maketitle

\begin{abstract}

In a spiking neural network (SNN), individual neurons operate autonomously and only communicate with
other neurons sparingly and asynchronously via spike signals. 
These characteristics render a massively parallel hardware implementation of SNN a potentially 
powerful computer, albeit a non von Neumann one. 
But can one guarantee that a SNN computer solves some important problems reliably?
In this paper, we formulate a mathematical model of one SNN
that can be configured for a sparse coding problem for feature extraction. With a moderate but 
well-defined assumption, we prove that the SNN indeed solves sparse coding.
To the best of our knowledge, this is the first rigorous result of this kind.

\end{abstract}

\section{Introduction}

A central question in computational neuroscience is to understand how complex
computations emerge from networks of neurons.
For neuroscientists, a key pursuit is to formulate neural network models that resemble
the researchers' understanding of physical neural activities and functionalities. Precise
mathematical definitions or analysis of such models is less important in comparison. 
For computer scientists, on the other hand, a key pursuit is often to devise new solvers
for specific computational problems. Understanding of neural activities
serves mainly as an inspiration for formulating neural network models; the actual model
adopted needs not be so much faithfully reflecting actual neural activities as
to be mathematically well defined and possesses provable properties such as stability
or convergence to the solution of the computational problem at hand.  

This paper's goal is that of a computer scientist. We formulate here two neural network
models that can provably solve a mixed $\ell_2$-$\ell_1$ optimization problem
(often called a LASSO problem). LASSO is a workhorse for sparse coding, a method
applicable across machine learning, signal processing, and statistics.
In this work, we provide a framework to rigorously establish the convergence of firing
rates in a spiking neural network to solutions corresponding to a LASSO problem.
This network model, namely the Spiking LCA, 
is first proposed in \cite{ShaperoRozellHasler13} to implement
the LCA model \cite{rozell2008sparse} using analog integrate-and-fire neuron circuit.
We will call the LCA model in~\cite{rozell2008sparse} the Analog LCA (A-LCA) for clarity. 
In the next section, we introduce the A-LCA model and its configurations for LASSO
and its constrained variant CLASSO. A-LCA is a form of Hopfield network, but the specific
(C)LASSO configurations render convergence difficult to establish. We will outline our recent
results that use a suitable generalization of the LaSalle principle to
show that A-LCA converges to (C)LASSO solutions.

In A-LCA, neurons communicate among themselves with real numbers (analog values) during
certain time intervals. In Spiking LCA (S-LCA), neurons communicate among themselves via
``spike'' (digital) signals that can be encoded with a single bit. Moreover, communication
occurs only at specific time instances. Consequently, S-LCA is much more communication 
efficient. Section~\ref{sec:snn} formulates S-LCA and other auxiliary variables such
as average soma currents and instantaneous spike rates. The section subsequently provides
a proof that the instantaneous rates converge to CLASSO solutions. This proof is built
upon the results we obtained for A-LCA and an assumption that a neuron's inter-spike
duration cannot be arbitrarily long unless it stops spiking altogether after a finite time.

Finally, we devise a numerical implementation of S-LCA and empirically demonstrate its convergence
to CLASSO solutions.
Our implementation also showcases the potential power of problem solving with spiking neurons in practice:
when an approximate implementation of S-LCA is ran on a conventional CPU,
it is able to converge to a solution with modest accuracy in a short amount of time.
The convergence is even faster than FISTA \cite{beck2009fast}, one of the fastest LASSO solvers.
This result suggests that a specialized spiking neuron hardware is promising,
as parallelism and sparse communications between neurons can be fully leveraged in such an
architecture.

\section{Sparse Coding by Analog LCA Neural Network} \label{sec:lca}

We formulate the sparse coding problem as follows. Given $N$ vectors in $\realsM$,
$\Phi = \left[\vvphi_1, \vvphi_2, \ldots, \vvphi_N\right]$, $N > M$, ($\Phi$ is usually called
a redundant---due to $N > M$---dictionary) and a vector $\vvs \in \realsM$ (consider $\vvs$
an input signal), try to code (approximate well) $\vvs$ as $\Phi \vva$ where $\vva\in\realsN$
contains as many zero entries as possible. 
Solving a sparse coding problem has attracted a tremendous amount of 
research effort~\cite{elad2010sparse}. One effective
way is to arrive at $\vva$ through solving the LASSO problem~\cite{tibshirani96} where
one minimizes the $\ell_2$ distance between $\vvs$ and $\Phi\vva$ with a $\ell_1$
regularization on the $\vva$ parameters. 
For reasons to be clear later on, we will consider this problem
with the additional requirement that $\vva$ be non-negative: $\vva\ge\vvzero$. 
We call this the CLASSO (C for constrained) problem:

\begin{equation}\label{eqn:CLASSO}
\operatorname{argmin}_{\vva\ge\vvzero}\;\;
\frac{1}{2}\Enorm{\vvs-\Phi\vva}^2 + \lambda\Onenorm{\vva}
\end{equation}

Rozell, et al., presented in~\cite{rozell2008sparse} the first neural network model aims at solving LASSO.
$N$ neurons are used to represent each of the $N$ dictionary atoms $\vvphi_i$. Each neuron receives
an input signal $b_i$ that serves to increase a ``potential'' value $u_i(t)$ that a neuron keeps
over time. When this potential is above a certain threshold, 
neuron-$i$ will send inhibitory signals that aim to reduce the potential values
of the list of receiving neurons with which neuron-$i$ ``competes.''
The authors called this kind of algorithms expressed in this neural network mechanism Locally Competitive Algorithms (LCAs).
In this paper, we call this as analog LCA (A-LCA).
Mathematically, an A-LCA can be described as a set of ordinary differential equations 
(a dynamical system) of the form
\begin{equation*}
\dot{u}_i(t) = b_i - u_i(t) - \sum_{j\neq i} w_{ij} \thres(u_j(t)), \quad i=1,2,\ldots,N.
\end{equation*}
The function $\thres$ is a thresholding (also known as an activation) function
that decides when and how an inhibition signal is sent. The coefficients $w_{ij}$ further
weigh the severity of each inhibition signal.
In this general form, A-LCA is an instantiation of the Hopfield network proposed in~\cite{hopfield82,hopfield84}.

Given a LASSO or CLASSO problem, A-LCA is configured by $b_i = \transpose{\vvphi}_i\,\vvs$,
$w_{ij} = \transpose{\vvphi}_i\,\vvphi_j$. For LASSO, the thresholding function
$\thres$ is set to $\thres = \threslasso$, and for CLASSO it is set to $\thres = \thresclasso$:
$\thresclasso(x)$ is defined as $0$ when $x\le \lambda$ and $x - \lambda$ when $x > \lambda$; 
and $\threslasso(x) \bydef \thresclasso(x) + \thresclasso(-x)$.
%
%
Note that if all the $\vvphi_i$s are normalized to $\transpose{\vvphi}_i \vvphi_i = 1$, then
the dynamical system in vector notation is
\begin{equation}\label{eqn:LCA}
\dot{\vvu} = \vvb - \vvu - (\transpose{\Phi}\Phi - I)\vva, \quad
\vva = \vvthres(\vvu).
\end{equation}
The vector function $\vvthres:\realsN\rightarrow\realsN$ simply applies the same scalar function
$\thres$ to each of the input vector's component. 
We say A-LCA solves (C)LASSO if a particular solution of the dynamical system
converges to a vector $\vvustar$ and that $\vvastar = \vvthres(\vvustar)$
is the optimal solution for (C)LASSO. This convergence phenomenon was demonstrated in~\cite{rozell2008sparse}.

LCA needs not be realized on a traditional computer via some classical numerical differential equation
solver; one can realize it using, for example, an analog circuit which may in fact be able to solve 
(C)LASSO faster or with less energy. From the point of view of establishing A-LCA as a robust
way to solve (C)LASSO, rigorous mathematical results on A-LCA's convergence is invaluable.
Furthermore, any convergence theory here will bound to have bearings on other neural network architectures, 
as we will see in Section~\ref{sec:snn}. 
Had the thresholding function $\thres$ in A-LCA be strictly increasing and unbounded above and below,
standard Lyapunov theory can be applied to establish convergence of the dynamical system. This
is already pointed out in Hopfield's early work for both graded neuron model~\cite{hopfield84}
and spiking neuron model~\cite{hopfield1995rapid}. Nevertheless, such an A-LCA does not
correspond to (C)LASSO where the thresholding functions are not strictly increasing. 
Furthermore, the CLASSO thresholding function is bounded below as well. While Rozell, et al.,
demonstrated some convergence phenomenon~\cite{rozell2008sparse}, it is in two 
later works~\cite{balavoine12,balavoine13} that Rozell and other colleagues attempted to complement the 
original work with convergence analysis and proofs.
Among other results, these works stated that for any particular A-LCA solution $\vvu(t)$,
$\vvthres(\vvu(t))$ with $\thres = \threslasso$ converges to a LASSO optimal solution. 
Unfortunately, as detailed in~\cite{tang2016lca}, there are major gaps in the related proofs
and thus the convergence claims are in doubt. Moreover, the case
of $\thres=\thresclasso$ for the CLASSO problem was not addressed.
In~\cite{tang2016lca}, one of our present authors established several convergence results
which we now summarize so as to support the development of Section~\ref{sec:snn}. 
The interested reader can refer to~\cite{tang2016lca} for complete details.

A-LCA is a dynamical system of the form $\dot{\vvu} = \vvF(\vvu)$, $\vvF:\realsN\rightarrow\realsN$.
In this case, the function $\vvF$ is defined as 
$\vvF(\vvx) = \vvb - \vvx - (\transpose{\Phi}\Phi - I)\vvthres(\vvx)$. Given any ``starting point''
$\ustart\in\realsN$, standard theory of ordinary differential equations shows that there is a unique
solution $\vvu(t)$ such that $\vvu(0)=\ustart$ and $\dot{\vvu}(t) = \vvF(\vvu(t))$ for all $t\ge 0$.
Solutions are also commonly called flows. The two key questions are (1) given some (or any) starting
point $\ustart$, whether and in what sense the flow $\vvu(t)$ converges, and (2) if so,
what relationships exist between the limiting process and the (C)LASSO solutions.

The LaSalle invariance principle~\cite{lasalle60} is a powerful tool to help answer the first question.
The gist of the principle is that if one can construct a function $V:\realsN\rightarrow\reals$ such
that it is non-increasing along any flow, then one can conclude that all flows must converge to a
special set\footnote{$\vvu(t)\tendsto\ccM$ if $\dist{\vvu(t)}{\ccM}\tendsto 0$ where
$\dist{\vvx}{\ccM}=\inf_{\vvy\in\ccM}\Enorm{\vvx-\vvy}$.}
$\ccM$ which is the largest 
positive invariant set\footnote{A set is positive invariant if any flow originated from the set stays
in that set forever.}
inside the set of points at which the Lie derivative of $V$ is zero. The crucial technical 
requirements on $V$ are that $V$ possesses continuous partial derivatives and
be radially unbounded\footnote{The function $V$ is radially unbounded if $|V(\vvu)|\tendsto\infty$
whenever $\Enorm{\vvu}\tendsto\infty$}.
Unfortunately, the natural choice of $V$ for A-LCA does not have continuous first partial
derivatives everywhere, and not radially unbounded in the case of CLASSO. Both failures
are due to the special form of $\vvthres$ with
$
V(\vvu) = (1/2)\Enorm{\vvs - \Phi\vvthres(\vvu)}^2 + \lambda\,\Onenorm{\vvthres(\vvu)}.
$
Based on a generalized version of LaSalle's principle proved in~\cite{tang2016lca}, we establish that
any A-LCA flow $\vvu(t)$ (LASSO or CLASSO) converges to $\ccM$, the largest positive invariant set
inside the ``stationary'' set  
$\ccS = \{\,\vvu \suchthat \hbox{$(\partial V/\partial u_n) F_n(\vvu) = 0$ 
whenever $|\thres(u_n)| > 0$.}\,\}$.

Having established $\vvu(t)\tendsto\ccM$, we further prove in~\cite{tang2016lca} that
$\ccM$ is in fact the inverse image under $\vvthres$ of the set $\ccC$ of optimal (C)LASSO solutions.
The proof is based on the KKT~\cite{Boyd} condition that characterizes $\ccC$ and
properties particular to A-LCA.

\begin{mythm}(A-LCA convergence results from~\cite{tang2016lca})
Given the A-LCA 
\begin{equation*}
\dot{\vvu} = \vvF(\vvu), \quad 
\vvF(\vvu) = \vvb - \vvu - (\transpose{\Phi}\Phi-I)\vvthres(\vvu).
\end{equation*}
$\vvthres$ is based on $\threslasso$ if one wants to solve LASSO and $\thresclasso$, CLASSO. 
Let $\ustart$ be an arbitrary starting point and $\vvu(t)$ be the corresponding flow.
The following hold:
\begin{enumerate}
\item Let $\ccC$ be the set of (C)LASSO optimal solutions and $\fixedregion = \inverse{\vvthres}(\ccC)$
be $\ccC$'s inverse image under the corresponding thresholding function $\vvthres$. Then 
any arbitrary flow $\vvu(t)$ always converges to the set $\fixedregion$.

\item Moreover, 
$\lim_{t\tendsto\infty} E(\vva(t)) = E^*$ where
$E^*$ is the optimal objective function value of (C)LASSO,
$E(\vva)=(1/2) \Enorm{\vvs-\Phi\vva}^2 + \lambda\Onenorm{\vva}$ and
$\vva(t) = \vvthres(\vvu(t))$.

\item Finally, when the (C)LASSO optimal solution $\vvastar$ is unique, then there is a unique
$\vvustar$ such that $\vvF(\vvustar) = \vvzero$. Furthermore $\vvu(t)\tendsto\vvustar$ and
$\vvthres(\vvu(t)) \tendsto \vvthres(\vvustar) = \vvastar$ as $t\tendsto\infty$.
\end{enumerate}
\end{mythm}

\section{Sparse Coding by Spiking LCA Neural Network} \label{sec:snn}

A-LCA is inherently communication efficient: Neuron-$i$ needs to communicate to others only when its
internal state $u_i(t)$ exceeds a threshold, namely $|\thres(u_i(t))| > 0$. In a sparse coding problem, 
it is expected that the internal state will eventually stay perpetually below the threshold for many
neurons. Nevertheless, for the entire duration during which a neuron's internal state is above threshold,
constant communication is required. Furthermore, the value to be sent to other neurons are real valued
(analog) in nature.
In this perspective, a spiking neural network (SNN) model holds the promise of even greater communication efficiency. 
In a typical SNN, various internal states of a neuron are also continually evolving. 
In contrast, however, communication in the form of a spike---that is one bit---is sent to other neurons only when
a certain internal state reaches a level (a firing threshold). This internal state is reset
right after the spiking event, thus cutting off communication immediately until the time when
the internal state is ``charged up'' enough.  Thus communication is necessary only once in
a certain time span and then a single bit of information carrier suffices.

While such a SNN admits mathematical descriptions~\cite{ShaperoRozellHasler13,deneve2013firingrate},
there is hitherto no rigorous results on the network's convergence behavior. In particular,
it is unclear how a SNN can be configured to solve specific problems with some guarantees. 
We present now a mathematical formulation of a SNN and a natural definition of instantaneous
spiking rate. Our main result is that under a moderate assumption,
the spiking rate converges to the CLASSO solution when the SNN is suitably configured. 
To the best of our knowledge, this is the first time a rigorous result of this kind is established.

In a SNN each of the $N$ neurons maintains, over time $t$, an internal soma current $\curr_i(t)$
configured to receive a constant input $b_i$ and an internal potential $\poten_i(t)$.
The potential is ``charged'' up according to $\poten_i(t)=\int_0^t(\curr_i-\currbias)$ where 
$\currbias\ge 0$ is a configured bias current. When $\poten_i(t)$ reaches a firing threshold $\potenfire$
at a time $\spiketime{i}{k}$, neuron-$i$ resets its potential to $\potenreset$ but simultaneously
fires an inhibitory signal to a preconfigured set of receptive neurons, neuron-$j$s, 
whose soma current will be diminished
according to a weighted exponential decay function: 
$\curr_j(t) \gets \curr_j(t) - w_{ji}\decay(t-\spiketime{i}{k})$,
where $\decay(t) = e^{-t}$ for $t \ge 0$ and zero otherwise. Let $\{\spiketime{i}{k}\}$ be the ordered time sequence
of when neuron-$i$ spikes and define $\spiketrain_i(t) = \sum_{k} \spike(t-\spiketime{i}{k})$, then
the soma current satisfies both the algebraic and differential equations below
(the operator $\conv$ denotes convolution):
\begin{equation}\label{eqn:current_DAE}
\curr_i(t) = \inputcurr_i - \sum_{j\neq i} w_{ij}(\decay \conv \spiketrain_j)(t), \quad
\dot{\curr}_i(t) = \inputcurr_i - \curr_i(t) - \sum_{j\neq i} w_{ij} \spiketrain_j(t).
\end{equation}
Equation~\ref{eqn:current_DAE} together with the definition of the spike trains $\spiketrain_i(t)$
describe our spiking LCA (S-LCA). 

An intuitive definition of spike rate of a neuron is clearly the number of spikes per unit time. 
Hence we define the instantaneous spiking rate $\spikerate_i(t)$ and average soma current
$\avgcurr_i(t)$ for neuron-$i$ as:
\begin{equation}\label{eqn:rates-def}
\spikerate_i(t) \bydef \frac{1}{t-\starttime}\int_{\starttime}^t \spiketrain_i(s)\,ds \quad
{\rm and}\quad
\avgcurr_i(t) \bydef \frac{1}{t-\starttime}\int_{\starttime}^t \curr_i(s)\,ds,
\quad\hbox{$\starttime\ge 0$ is a parameter.} 
\end{equation}
Apply the operator
$(t-\starttime)^{-1}\int_{\starttime}^t\;ds$
to the differential equation portion in~(\ref{eqn:current_DAE}), using also 
the relationship $\dot{\avgcurr}_i(t) = (\curr_i(t)-\avgcurr_i(t))/(t-\starttime)$, 
and we obtain 
\begin{equation}\label{eqn:rates-DE}
\dot{\avgcurr}_i(t) = \inputcurr_i - \avgcurr_i(t) - \sum_{j\neq i} w_{ij} \spikerate_j(t) -
(\avgcurr_i(t) - \avgcurr_i(\starttime))/(t-\starttime).
\end{equation}

Consider now a CLASSO problem where the dictionary atoms are non-negative and normalized to unit Euclidean norm.
Configure S-LCA with $\currbias$ and $w_{ij} = \transpose{\vvphi}_i\vvphi_j$ from Equation~\ref{eqn:CLASSO},
and set $\potenfire \gets 1$, $\potenreset \gets 0$.
So configured, it can be shown that the soma currents' magnitudes (and thus that of the average currents as well)
are bounded: there is a $B$ such that $|\curr_i(t)|, |\avgcurr_i(t)| \le B$ for all $i = 1, 2, \ldots, N$ and all 
$t > \starttime$. Consequently, 
\begin{equation}\label{eqn:udot_goesto_zero}
\lim_{t\tendsto\infty} \dot{\avgcurr}_i(t) = \lim_{t\tendsto\infty} (\curr_i(t)-\avgcurr_i(t))/(t-\starttime) = 0, 
\quad \hbox{for $i = 1,2,\ldots,N$}.
\end{equation}
The following relationship between $\avgcurr_i(t)$ and $\spikerate_i(t)$ is crucial:
\begin{equation}\label{eqn:u_and_a}
\avgcurr_i(t) - \currbias =
\frac{1}{t-\starttime} \int_{\starttime}^{\spiketime{i}{k}} (\curr_i-\currbias) +
\frac{1}{t-\starttime} \int_{\spiketime{i}{k}}^t (\curr_i-\currbias) 
= \spikerate_i(t) + \poten_i(t)/(t-\starttime).
\end{equation}
From this equation and a moderate assumption that inter-spike duration 
$\spiketime{i}{k+1} - \spiketime{i}{k}$ cannot
be arbitrarily long unless neuron-$i$ stops spiking altogether, one can prove that
\begin{equation}\label{eqn:a_and_Tu}
\thresclasso(\avgcurr_i(t)) - \spikerate_i(t) \tendsto 0 \quad \mbox{as $t \tendsto \infty$}.
\end{equation}
The complete proof for this result is left in the Appendix.


We can derive convergence of S-LCA as follows. Since the average soma currents 
are bounded, Bolzano-Weierstrass theorem shows that 
$\vvu(t) \bydef \transpose{[\avgcurr_1(t),\avgcurr_2(t),\ldots,\avgcurr_N(t)]}$ has at least one limit point,
that is, there is a point $\vvustar\in\realsN$ and a time sequence $t_1 < t_2 < \cdots$, $t_k\tendsto\infty$
such that $\vvu(t_k) \tendsto \vvustar$ as $k\tendsto\infty$. By Equation~\ref{eqn:a_and_Tu},
$\vvthres(\vvu(t_k)) \tendsto \vvthres(\vvustar) \bydef \vvastar$. By Equations~\ref{eqn:rates-DE}
and~\ref{eqn:udot_goesto_zero}, we must therefore have
\begin{equation}\label{eqn:snn_fixedpoint}
\vvzero = \vvb - \vvustar - (\transpose{\Phi}\Phi - I)\vvastar.
\end{equation}
Since S-LCA is configured for a CLASSO problem, the limit $\vvustar$ is in fact a fixed point
of A-LCA, which is unique whenever CLASSO's solution is. In this case, the limit point
of the average currents $\{\vvu(t) \suchthat t \ge 0 \}$ is unique and thus indeed
we must have $\vvu(t)\tendsto\vvustar$ and $\vva(t) \tendsto \vvastar$, the CLASSO solution.

\section{Numerical Simulations}

To simulate the dynamics of S-LCA on a conventional CPU, one can precisely solve the continuous-time 
spiking network formulation by tracking the order of firing neurons.
In between consecutive spiking events, the internal variables, $v_i(t)$ and $\mu_i(t)$, of each neuron follow simple
differential equations and permit closed-form solutions.
This method, however, is likely to be slow that it requires a global coordinator that looks ahead into the future to determine the next firing neuron.
For efficiency, we instead take an approximate approach that evolves the network state in constant-sized discrete time steps.
At every step, the internal variables of each neuron are updated and a firing event is triggered if the potential 
exceeds the firing threshold. 
The simplicity of this approach admits parallel implementations and is suitable for specialized hardware designs. 
Nevertheless, this constant-sized-time-step approach
introduces errors in spike timings: the time that a neuron sends out a spike
may be delayed by up to a time step. 
As we will see in this section, the timing error is the major factor that limits the accuracy of 
the solutions from spiking networks. However, such an efficiency-accuracy trade-off 
may in fact be desirable for certain applications such as those in machine learning.

\subsection{Illustration of SNN dynamics}

\begin{figure}[t]
\centering
\setlength\tabcolsep{2.0pt}
  \begin{tabular}{cc}
    \begin{tabular}{c}
      \includegraphics[scale=0.55]{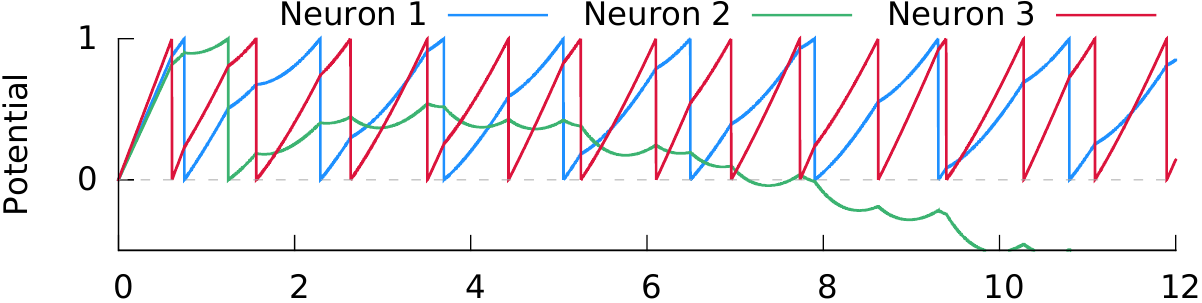} \\ (a) \\
      \includegraphics[scale=0.55]{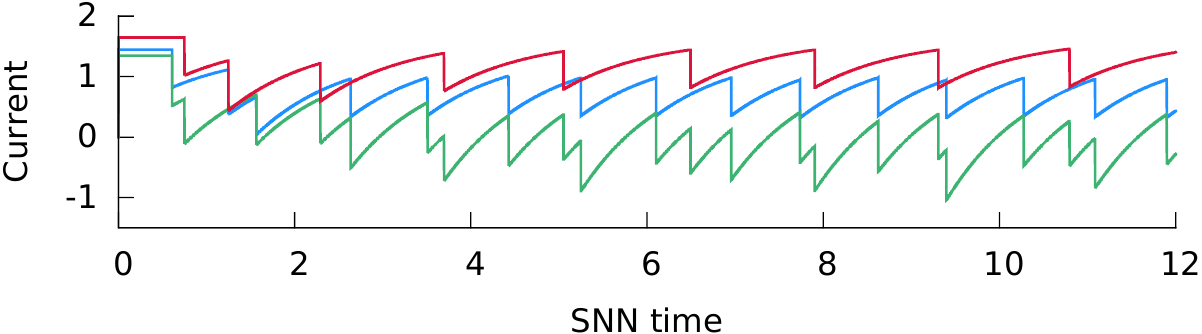} \\ (b) 
    \end{tabular}
    & 
    \begin{tabular}{c}
      \includegraphics[scale=0.55]{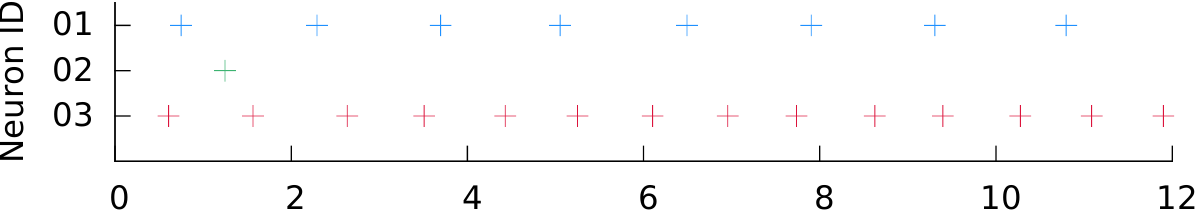}\\ (c) \\
      \includegraphics[scale=0.55]{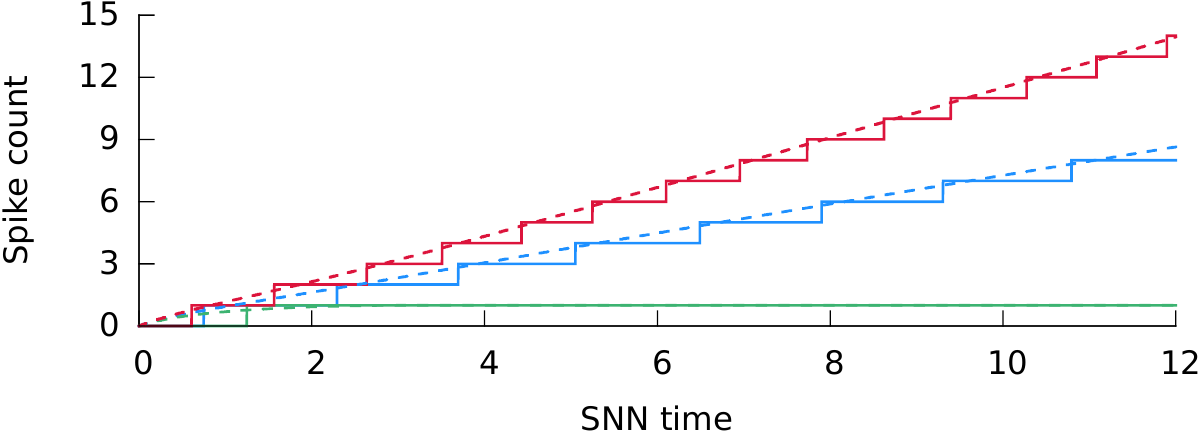} \\ (d)
    \end{tabular}
  \end{tabular}
  \caption{
Detail dynamics of a simple 3-neuron spiking network:
In the beginning before any neuron fires, the membrane potentials (see 1-a) 
of the neurons grow linearly with a rate determined by the initial soma currents (see 1-b). 
This continues until Neuron 3 becomes the first to reach the firing threshold; 
an inhibitory spike is sent to Neurons 1 and 2, causing immediate drops in their soma currents.
Consequently, the growths of Neurons 1 and 2's membrane potentials slow down, and 
the neurons' instantaneous spike rates decrease.
The pattern of membrane integration, spike, and mutual inhibition repeats; 
the network rapidly moves into a steady state where stable firing rates can be observed. 
The convergent firing rates yield the CLASSO optimal solution of 
$[0.684, 0, 1.217]$ (this solution is also verified by running the LARS
algorithm \protect\cite{efron2004least}).
The four sub-figures:
(a) Evolution of membrane potential. (b) Evolution of soma current. 
(c) Spike raster plot. (d) Solid lines are the cumulative spike count of each neuron, and
dashed line depicts the value of $\int_{0}^{t}T_\lambda(u_i(s))ds$ in 
the corresponding A-LCA. The close approximation indicates a strong tie between
the two formulations.}
\label{fig:dynamics}
\end{figure}

We solve a simple CLASSO problem: $\min_{\vva} {1\over2}\Enorm{\vvs-\Phi\vva}^2 + \lambda\,\Onenorm{\vva}$ 
subject to $\vva\ge\vvzero$, where
\begin{equation*}
\vvs = 
\begin{bmatrix}
0.5\\ 
1\\ 
1.5
\end{bmatrix}
,\quad 
\Phi = [\vvphi_1 \, \vvphi_2 \, \vvphi_3] = 
\begin{bmatrix}
0.3313 & 0.8148 & 0.4364 \\ 
0.8835 & 0.3621 & 0.2182 \\ 
0.3313 & 0.4527 & 0.8729
\end{bmatrix}
,\quad
\lambda = 0.1.
\end{equation*}
We use a 3-neuron network configured with $b_i = \transpose{\vvphi}_i\vvs$, 
$w_{ij} = \transpose{\vvphi}_i\vvphi_j$, the bias current as $\lambda=0.1$ and
firing threshold set to 1.
Figure \ref{fig:dynamics} details the dynamics of this simple 3-neuron spiking network.
It can be seen from this simple example that the network only needs very few spike exchanges for it to converge.
In particular, a weak neuron, such as Neuron 2, is quickly rendered inactive by inhibitory
spike signals from competing neurons.
This raises an important question: \textit{how many spikes are in the network?}
We do not find this question easy to answer theoretically.
However, empirically we see the number of spikes in $\spiketrain_i(t)$ in S-LCA can be approximated 
from the state variable $u_i$ in A-LCA, that is the $u_i(s)$ 
in Equation~\ref{eq:slca_num_spikes} below are solutions to Equation~\ref{eqn:LCA}:
\begin{equation}
  \int_{0}^{t}\spiketrain_i(s)ds \approx \int_{0}^{t} \thresclasso(u_i(s))ds, \quad i=1,2,\ldots,N.
\label{eq:slca_num_spikes}
\end{equation}
Figure \ref{fig:dynamics}(d) shows the close approximation of spike counts using (\ref{eq:slca_num_spikes}) in the example.
We observe that such approximation consistently holds in large-scale problems, suggesting a strong
tie between S-LCA and A-LCA.
Since in an A-LCA configured for a sparse coding problem, we expect 
$\thresclasso(u_i(t))$ for most $i$'s to converge to zero,
(\ref{eq:slca_num_spikes}) suggests that the total spike count in S-LCA is small.

\subsection{Convergence of spiking neural networks}

\begin{figure}[t]
\centering	
  \begin{tabular}{cc}
    \begin{tabular}{c}
      \includegraphics[scale=0.52]{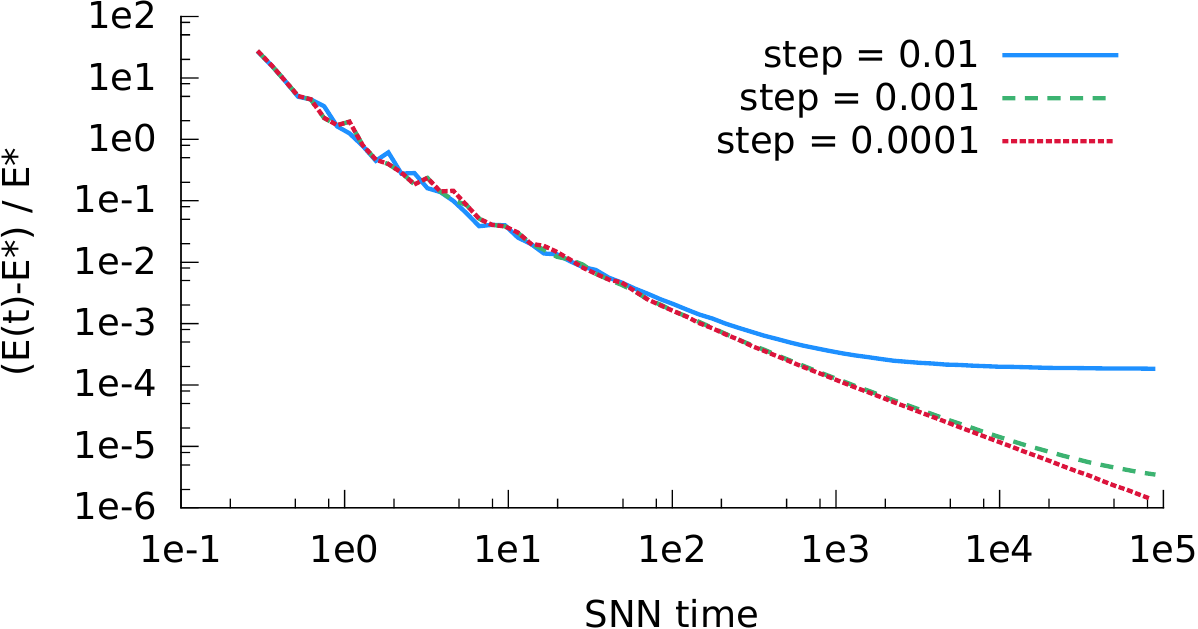} \\ (a) 
    \end{tabular}
    &
    \begin{tabular}{c}
      \includegraphics[scale=0.52]{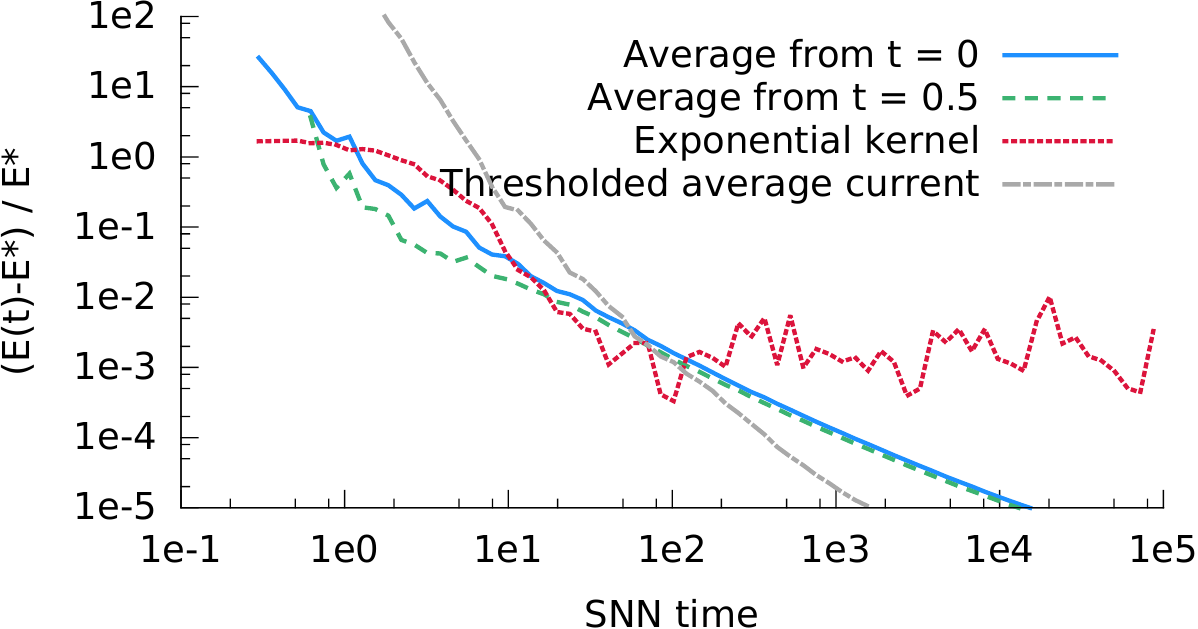} \\ (b) 
    \end{tabular}
  \end{tabular}
  \caption{(a) The convergence of a 400-neuron spiking network to a CLASSO solution. 
   (b) Comparing the convergence of different formulations
    to read out solutions from a spiking neural network. 
Using a positive $t_0$ for Equation~\ref{eqn:rates-def} gives 
the fastest initial convergence, while using the thresholded average current
reaches the highest accuracy the quickest.
Despite a lack of theoretical guarantee, the exponential kernel method yields
an acceptable, though less accurate, solution. This kernel is easy to implement in hardware 
and thus attractive when a SNN ``computer'' is to be built.
}
  \label{fig:convergence}
\end{figure}

We use a larger 400-neuron spiking network to empirically examine the convergence of spike rates to CLASSO solution.
The neural network is configured to perform feature extraction from a 8$\times$8 image patch, using a
400-atom dictionary learned from other image 
datasets.\footnote{The input has 128 dimensions by splitting the image into positive and negative channels.}
With the chosen $\lambda$, the optimal solution has 8 non-zeros entries.
Figure \ref{fig:convergence}(a) shows the convergence of the objective function value in the spiking network solution, comparing to the true
optimal objective value obtained from a conventional CLASSO solver. 
Indeed, 
with a small step size, the spiking network converges to a solution very close
to the true optimum.

The relationships among step size, solution accuracy and total computation cost are noteworthy.
Figure~\ref{fig:convergence}(a) shows that increasing the step size from $10^{-3}$ to $10^{-2}$ sacrifices
two digits of accuracy in the computed $E^*$. The total computation cost is reduced by a factor of $10^3$:
It takes $10^2$ times fewer time units to converge, and each time unit requires $10$ times fewer iterations.
This multiplication effect on cost savings is highly desirable in
applications such as machine learning where accuracy is not paramount.   
We note that a large-step-size configuration is also suitable for problems whose solutions are sparse:
The total number of spikes are fewer and thus total timing errors are correspondingly fewer.

There are several ways to ``read out'' a SNN solution.
Most rigidly, we can adhere to $\spikerate_i(t)$ in Equation~\ref{eqn:rates-def} with $\starttime = 0$.
In practice, picking some $\starttime > 0$ is better when we expect a sparse solution:
The resulting $\spikerate_i(t)$ will be identically zero for those neurons that only spike
before time $\starttime$.
Because $\thresclasso(\avgcurr_i(t)) - \spikerate_i(t) \tendsto 0$ (Equation~\ref{eqn:a_and_Tu}),
another alternative is to use $\thresclasso(\avgcurr_i(t))$ as the solution, which is 
more likely to deliver a truly sparse solution. 
Finally, one can change $\spikerate_i(t)$'s definition 
to $\tau^{-1}\int_{0}^{t} e^{-\frac{t-s}{\tau}} \spiketrain_i(s)ds$, 
so that the impact of the spikes in the past decays quickly. 
Figure~\ref{fig:convergence}(b) illustrates these different ``read out'' methods
and shows that the exponential kernel is as effective empirically, although we
must point out that the previous mathematical convergence analysis is no
longer applicable in this case.

\subsection{CPU benchmark of a spiking network implementation}

Our earlier discussions suggest that the spiking network can solve CLASSO using very few spikes.
This property has important implications to a SNN's computational efficiency.
The computation cost of a $N$-neuron spiking network has two components:
neuron states update and spiking events update.
Neuron states update includes updating the internal potential and current values of every neuron,
and thus incurs an $O(N)$ cost at every time step. 
The cost of spiking events update is proportional
to $N$ times the average number of inter-neuron connections
because a spiking neuron updates the soma currents of those
neurons to which it connects.
Thus this cost can be as high as $O(N^2)$ (for networks with all-to-all connectivity,
such as in the two previous examples) or  
as low as $O(N)$ (for networks with only local connectivity, such as in the example below).
Nevertheless, spiking-event cost is incurred only when there is a spike, which may 
happen far fewer than once per time step. 
In practice we observe that computation time is usually dominated by neuron-states update,
corroborating the general belief that spiking events are relatively rare, making
spiking networks communication efficient. 

\begin{figure}[t]
\centering
\setlength\tabcolsep{1.5pt}
  \begin{tabular}{ccc}
    \begin{tabular}{c}
      \includegraphics[scale=0.38]{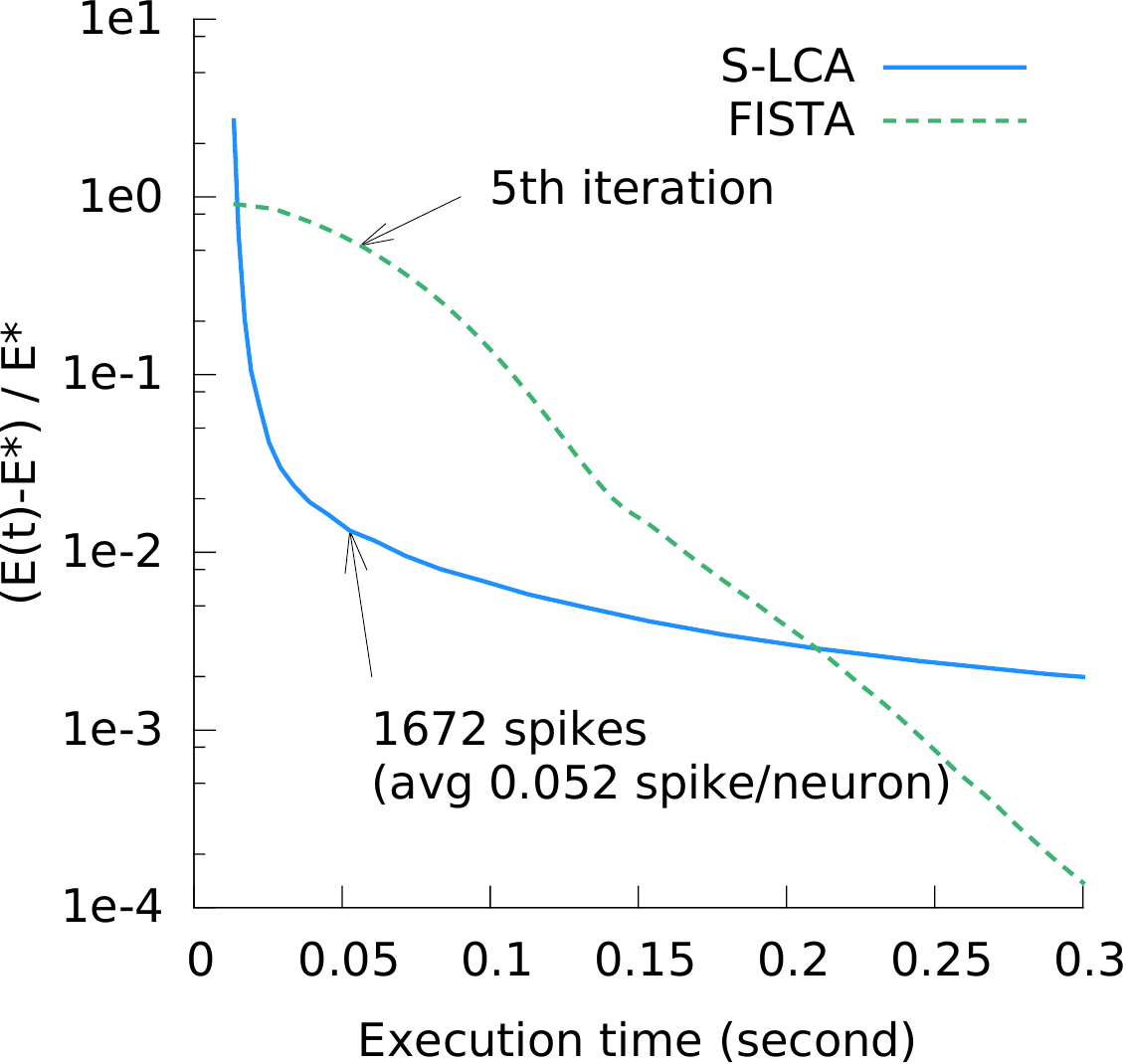} \\ 
      (a) 52$\times$52 image
    \end{tabular}
    &
    \begin{tabular}{c}
      \includegraphics[scale=0.42]{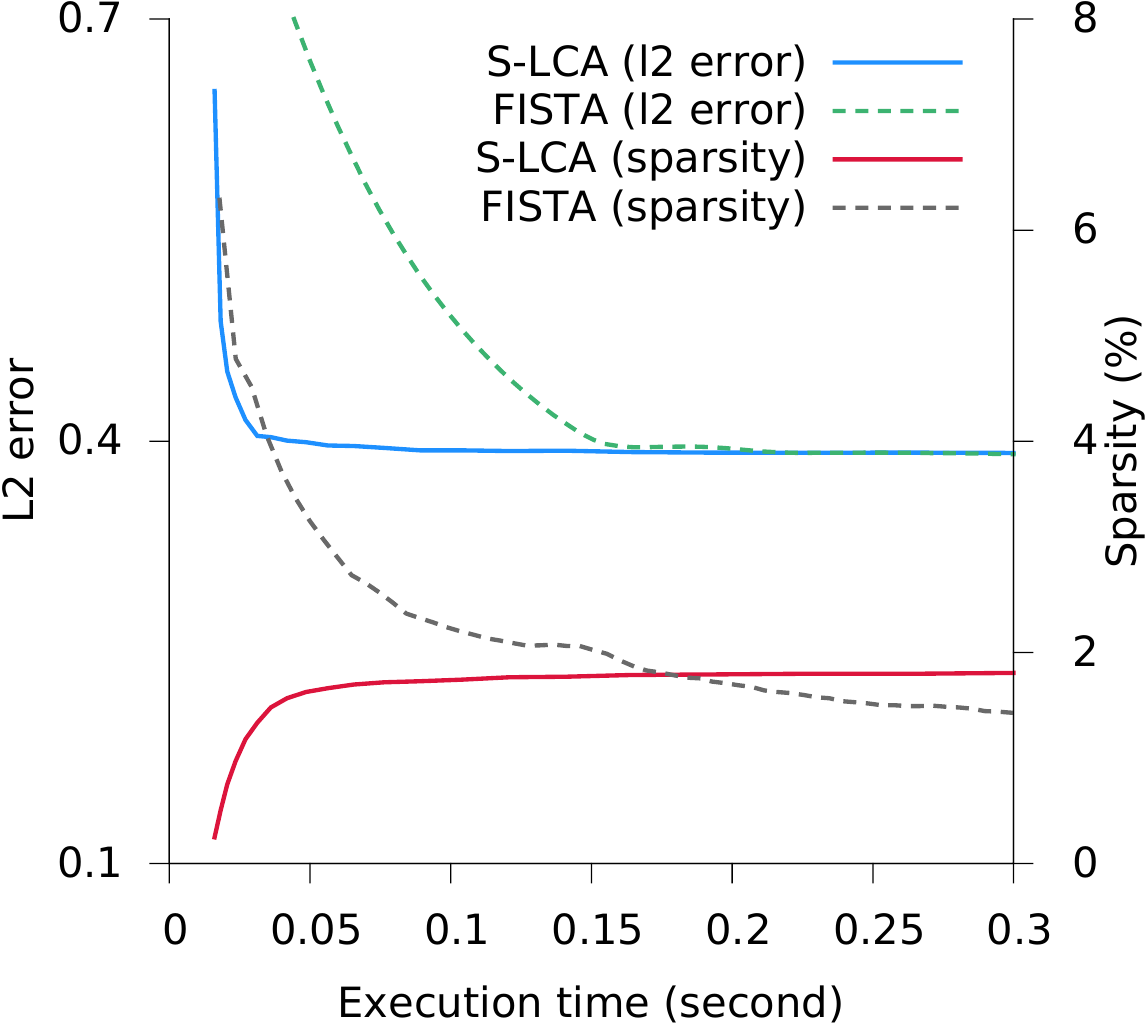} \\
      (b) Breakdown of (a)
    \end{tabular}
    &    
    \begin{tabular}{c}
      \includegraphics[scale=0.38]{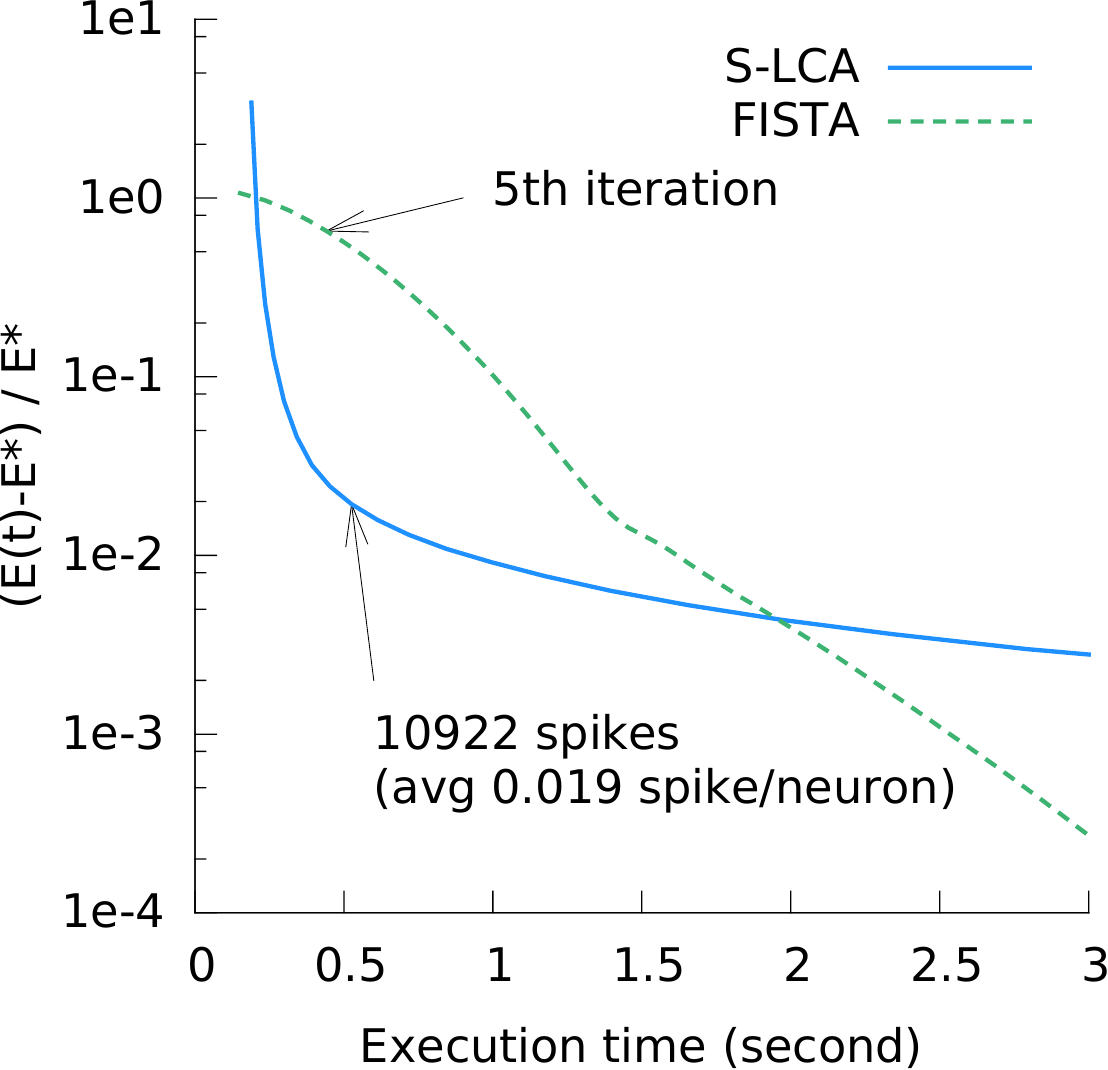} \\ 
      (c) 208$\times$208 image
    \end{tabular}
  \end{tabular}
  \caption{CPU execution time for spiking neural networks, with a step size of 0.01. 
    There are 32,256 unknowns in the 52$\times$52 image case shown in (a), and 582,624 unknowns
    in the 208$\times$208 image case shown in (c). (b) shows the breakdown of the objective function in the
        52$\times$52 image experiment. The $\ell_2$ error is defined as 
        $\frac{\Enorm{\vvs-\Phi\vva}}{\Enorm{\vvs}}$,
        and the sparsity is the percentage of entries with values greater than 0.01.
        Note that the spiking network finds the optimal solution by gradually increasing the
        sparsity, rather than decreasing as in FISTA. This results in the spare spiking activities
        of the neurons.
      }
  \label{fig:benchmark}
\end{figure}

We report the execution time of simulating the spiking neural network on a conventional CPU, 
and compare the convergence time with FISTA \cite{beck2009fast}, one of the fastest LASSO solvers.
We solve a convolutional sparse coding problem \cite{zeiler2010deconvolutional}
on a 52x52 image and a 208x208 image.\footnote{We use 8$\times$8 patches, a stride of 4, and 
a 128$\times$224 dictionary.} 
The experiments are ran on 2.3GHz Intel\textsuperscript{\textregistered} Xeon\textsuperscript{\textregistered} CPU E5-2699 using a single core.
SIMD is enabled to exploit the intrinsic parallelism of neural network and matrix operations.
As shown in Figure \ref{fig:benchmark}, the spiking network delivers much faster early convergence than FISTA, 
despite its solution accuracy plateauing due to spike timing errors.
The convergence trends in both figures are similar, demonstrating that spiking 
networks can solve problems of various sizes.
The fast convergence of spiking networks can be attributed to their ability to fully exploit the
sparsity in solutions to reduce the spike counts.
The fine-grain asynchronous communication can quickly
suppress most neurons from firing.
In FISTA or in any other conventional solvers, communications between variables is similarly needed, but is 
realized through matrix-vector multiplications performed in an iteration-to-iteration basis.
The only way to exploit sparsity is to avoid computations involving variables that have gone to zero
during one iteration.
A comparison of how the sparsity in solutions evolves in S-LCA and FISTA can be found in Figure \ref{fig:benchmark}(b).

\section{Discussion}

Our work is closely related to the recent progress on optimality-driven 
balanced network~\cite{deneve2016efficient,deneve2013firingrate,boerlin2013predictive}.
The SNN model in~\cite{deneve2013firingrate,boerlin2013predictive} differs slightly from ours in that only
one internal state is used in the former. Using our language here, 
neuron-$i$'s spike is generated by $\curr_i(t)$ reaching a threshold and
not by $\poten_i(t)$, whose role is eliminated altogether.
Despite the differences in the details of neuron models,
spikes in both networks occur from a competitive process between neurons,
and serve to minimize a network-level energy function.
This work furthers the understanding of the convergence property in such spiking networks.
Additionally, it is argued that in a tightly balanced excitatory/inhibitory network,
spike codes are highly efficient that
each spike is precisely timed to keep the network in optimality.
This work provides evidence of the high coding efficiency even before the network settles into 
steady-state.
By utilizing well-timed spikes, the neurons are able to collectively solve optimization problems 
with minimum communications.
We demonstrate that this insight can be translated into practical value 
through an approximate implementation on conventional CPU.

We observe that mathematical rigor was not a focus in~\cite{deneve2013firingrate}:
The statement that in a tightly balanced network the potential converges to zero is
problematic when taken literally as all spiking events will eventually cease
in that case. The stationary points of the loss function 
(Equation 6 in~\cite{deneve2013firingrate}) are no longer necessarily the
stationary points when the firing rates are constrained to be non-negative.
The general KKT condition has to be used in this situation. The condition
$E(\mbox{no spike}) > E(\mbox{spike})$ does not affect the behavior of
the loss function in between spikes. In essence, there is no guarantee
that the trajectory of the $\vvr(t)$ variable generated by the SNN is descending
the loss function, that is, $\frac{d}{dt}E(\vvr(t)) \le 0$.

Our SNN formulation and the established convergence properties can be easily extended to incorporate
an additional $\ell_2$-penalty term, the so-called elastic-net problem \cite{ZouHastie05}
\begin{equation}\label{eqn:elastic-net}
  \operatorname{argmin}_{\vva\ge\vvzero}\;\;
  \frac{1}{2}\Enorm{\vvs-\Phi\vva}^2 + \lambda_1\Onenorm{\vva} +\lambda_2\Enorm{\vva}^2
\end{equation}
The elastic-net formulation can be handled by modifying the slope of the activation function $T$ in A-LCA
as follows
\begin{equation*}
  \thresclasso(x) \bydef \left\{
    \begin{array}{l l}
      0          & \mbox{if $x \le \lambda_1$} \\
      \frac{x-\lambda_1}{2\lambda_2+1}  & \mbox{if $x > \lambda_1$}
    \end{array}\right., \quad
    \threslasso(x) \bydef \thresclasso(x) + \thresclasso(-x).
\end{equation*}
In S-LCA, this corresponds to setting the bias current to $\lambda_1$
and modifying the firing thresholds $\potenfire$ of the neurons to $2\lambda_2+1$.

There are several other works studying the computation of sparse representations using spiking neurons.
Zylberberg et al. \cite{zylberberg2011sparse} show the emergence of sparse representations through local rules, but do not provide a network-level 
energy function.
Hu et al. \cite{hu2012network} derive a spiking network formulation that minimizes a modified time-varying LASSO objective.
Shapero et al. \cite{ShaperoRozellHasler13,shapero2014optimal} are the first to propose the S-LCA formulation, but yet to provide an in-depth analysis.
We believe the S-LCA formulation can be a powerful primitive in future spiking network research. 

The computational power of spikes enables new opportunities in future computer architecture designs.
The spike-driven computational paradigm motivates an architecture composed of massively parallel computation units.
Unlike the von Neumann architecture, the infrequent but dispersed communication pattern between the units 
suggests a decentralized design where memory should be placed close to the compute, and communication
can be realized through dedicated routing fabrics.
Such designs have the potential to accelerate computations without breaking the energy-density limit.

\newpage
\appendix
\appendixpage

\section{Governing Algebraic and Differential Equations}~\label{sec:equations}

Consider a neural networking consisting of $N$ neurons. The only independent variables
are the $N$ soma currents $\curr_i(t)$ for $i=1,2,\ldots,N$. There are another $N$ variables
of potentials $\poten_i(t)$ which are depedent on the currents to be described momentarily.
Consider the following configurations. Each neron receives a positive constant input current
$\inputcurr_i$. A nonnegative current bias $\currbias$ and a positive potential 
threshold $\potenthres$ are set a priori. At any given
time $t_0$ such that $\poten_i(t_0) < \potenthres$, the potential evolves according to
$$
\poten_i(t) = \int_{t_0}^t (\curr_i(s) - \currbias)\,ds
$$
until the time $\spiketime{i}{k} > t_0$ when $\poten_i(t) = \potenthres$. 
At this time, a spike signal is sent from neuron-$i$ to all the nerons that are connected to it,
weighted by a set of pre-configured weights $w_{j,i}$. The potential $\poten_i(t)$ is reset to
zero immediately afterwards. That is, for $t > \spiketime{i}{k}$ but before the next spike 
is generated,
$$
\poten_i(t) = \int_{\spiketime{i}{k}}^t (\curr_i(s) - \currbias)\,ds.
$$
Moreover, for any consecutive spike times $\spt{i}{k}$ and $\spt{i}{k+1}$,
$$
\int_{\spt{i}{k}}^{\spt{i}{k+1}} (\curr_i(s)-\currbias)\,ds = \potenthres.
$$
Finally, when neuron-$i$ receives a spike from neuron-$j$ at time $\spiketime{j}{k}$ with a
weight $w_{i,j}$, the soma current $\curr_i(t)$ is changed by the additive signal
$-w_{i,j} \decay(t-\spiketime{j}{k})$ where
$$
\decay(t) = H(t) \expdecay{t},
$$
$H(t)$ being the Heaviside function that is 1 for $t \ge 0$ and 0 otherwise. The sign convention
used here means that a positive $w_{i,j}$ means that a spike from neuron-$j$ always tries to
inhibit neuron-$i$.

\VS
Suppose the initial potentials $\poten_i(0)$ are all set to be below the spiking threshold
$\potenthres$, then the dynamics of the system can be succintly described by the set of algebraic
equations

\begin{equation}\label{eqn:currents-AE}
\curr_i(t) = \inputcurr_i - \sum_{j\neq i} w_{i,j} \,(\decay\conv\spiketrain_j)(t),
\quad i=1,2,\ldots,N  \tag{AE}
\end{equation}
where $\conv$ is the convolution operator and $\spiketrain_j(t)$ is the sequence of
spikes 
$$
\spiketrain_j(t) = \sum_{k} \dirac(t-\spiketime{j}{k}),
$$
$\dirac(t)$ being the Dirac delta function.
The spike times are determined in turn by the evolution of the soma currents that
govern the evolutions of the potentials.

\VS
One can also express the algebraic equations~\ref{eqn:currents-AE} as a set of differential equations.
Note that the Heaviside function can be expressed as $H(t) = \int_{-\infty}^t \dirac(s)\,ds$.
Hence
\begin{eqnarray*}
\frac{d}{dt} \decay(t) 
& = & \frac{d}{dt} \left( \expdecay{t}\; \int_{-\infty}^t \dirac(s)\,ds \right) \\
& = & -\frac{1}{\expwidth}\,\decay(t) + \dirac(t).
\end{eqnarray*}
Thus, differentiating Equation~\ref{eqn:currents-AE} yields
\begin{equation}\label{eqn:currents-DE}
\dot{\curr}_i(t) = 
\frac{1}{\expwidth}\left(\inputcurr_i - \curr_i(t)\right) - \sum_{j\neq i} w_{i,j} \spiketrain_j(t).
\tag{DE}
\end{equation}

\VS
Note that Equations~\ref{eqn:currents-AE} and~\ref{eqn:currents-DE} are given in terms of the
spike trains $\spiketrain_j(t)$ that are governed in turn by the soma currents themselves as 
well as the configuartions of initial potentials, the spiking threshold $\potenthres$ and
bias current $\currbias$.

\section{Defining Spike Rates and Average Currents}\label{sec:spike-rates}

Suppose the system of spiking neurons are initialized with sub-threshold potentials, that is,
$\poten_i(0) < \potenthres$ for all $i=1,2,\ldots,N$. Thus at least for finite time after 0,
all soma currents remain constant at $\inputcurr_i$ and that no neurons will generate
any spikes. Furthermore, consider for now that $w_{i,j} \ge 0$ for all $i, j$. That is, only
inhibitory signals are present.
Let the spike times for each neuron
$i$ be $0 < t_{i,1} < t_{i,2} < \cdots$. This sequence could be empty, finite, or infinite.
It is empty if the potential $\poten_i(t)$ never reaches the threshold. It is finite
if the neuron stop spiking from a certain time onwards. We will define the spike rate,
$\spikerate_i(t)$, and average current, $\avgcurr_i(t)$, for each neuron as follows.
$$
\spikerate_i(t) \bydef \left\{
\begin{array}{l l}
\frac{1}{t}\int_0^t \spiketrain_i(s)\,ds  & t > 0, \\
0                                         & t = 0
\end{array}
\right. ,
$$
and
$$
\avgcurr_i(t) \bydef \left\{
\begin{array}{l l}
\frac{1}{t}\int_0^t \curr_i(s)\,ds  & t > 0, \\
\inputcurr_i                        & t = 0
\end{array}
\right. .
$$

\VS
With these definitions, the section presents the following results.
\begin{itemize}
\item The inhibition assumption leads to the fact that all the soma currents are bounded above.
      This in turns shows that none of the neurons can spike arbitrarily rapidly.
\item The fact that neurons cannot spike arbitrarily rapidly implies the soma currents are
      bounded from below as well.
\item The main assumption needed (that is, something cannot be proved at this point) is that
      if a neron spikes infinitely often, then the duration between consecutive spikes cannot
      be arbitrarily long. 
\item Using this assumption and previous established properties, one can prove an important 
      relationship between the spike rate and average current in terms of the familiar
      thresholding function $\thres$
\end{itemize}

\begin{myprop}\label{prop:curr-spikerate-bounds}
There exists bounds $B_-$ and $B_+$ such that $\curr_i(t)\in [B_-,B_+]$ for all $i$ and $t\ge 0$.
With the convention that $t_{i,0} \bydef 0$, then there is a positive value $R > 0$ such that
$t_{i,k+1} - t_{i,k} \ge 1/R$ for all $i = 1, 2, \ldots, N$ and $k \ge 0$, whenever these
values exist.
\end{myprop}
\begin{myproof}
Because all spike signals are inhibitory, clearly from Equation~\ref{eqn:currents-AE}, we have
$\curr_i(t) \le \inputcurr_i$ for all $t\ge 0$. Thus, defining $B_+ \bydef \max_i b_i$ leads
to $\curr_i(t) \le B_+$ for all $i$ and $t\ge 0$.

\VS
Given any two consecutive $t_{i,k}$ and $t_{i,k+1}$ that exist,
\begin{eqnarray*}
\potenthres 
 & = & \poten_i(t^+_{i,k}) + \int_{t_{i,k}}^{t_{i,k+1}} (\curr_i(s)-\currbias)\,ds  \\
 & \le & \poten_i(t^+_{i,k}) + (t_{i,k+1} - t_{i,k})(B_+ - \currbias).
\end{eqnarray*}
Note that $\poten_i(t^+_{i,k}) = 0$ if $k \ge 1$. For the special case when $k=0$, 
this value is $\poten_i(0) < \potenthres$. Hence
$$
t_{i,k+1} - t_{i,k} \ge \min\left\{ \min_i \{\potenthres-\poten_i(0)\}, \potenthres \right\}\, 
                           (B_+ - \currbias)^{-1}.
$$
Thus there is a $R > 0$ so that $t_{i,k+1} - t_{i,k} \ge 1/R$ whenever these two spike times exist.

\VS
Finally, because of duration between spikes cannot be arbitrarily small, it is easy to see that
$$
\gamma \bydef
\sum_{\ell=0}^\infty e^{\frac{-\ell}{R\expwidth}} \ge
(\decay\conv\spiketrain)(t). 
$$
Therefore,
$$
B_- \bydef
\min\{ -\gamma\,\sum_{j\ne i} w_{i,j} \} \le \curr_i(t)
$$
for all $i=1,2,\ldots,N$ and $t\ge 0$. So indeed, there are $B_-$ and $B_+$ such that
$\curr_i(t) \in [B_-,B_+]$ for all $i$ and $t\ge 0$.
\qed
\end{myproof}

\VS
Proposition~\ref{prop:curr-spikerate-bounds} shows that among other things, there is a lower bound
of the duration of consecutive spikes. The following is an assumption. 

\begin{myassumption}\label{assume:spike-duration-upper-bound}
Assume that there is a positive number $r > 0$ such that whenever the numbers
$t_{i,k}$ and $t_{i,k+1}$ exist, $t_{i,k+1} - t_{i,k} \le 1/r$.
\end{myassumption}
In simple words, this assumption says that unless a neuron stop spiking althogether after
a certain time, the duration between consecutive spike cannot become arbitrarily long.
With this assumption and the results in Proposition~\ref{prop:curr-spikerate-bounds},
the following important relationship between $\avgcurr(t)$ and $\spikerate(t)$ can be
established.

\begin{mythm}\label{thm:a-and-u}
Let $\thres(x)$ be the thresholding function where $\thres(x) = 0$ for $x \le \currbias$,
and $\thres(x) = x - \currbias$ for $x > \currbias$.
For each neuron $i$, there is a function $\Delta_i(t)$ such that
$$
\thres(\avgcurr_i(t)) = \spikerate_i(t)\,\potenthres + \Delta_i(t)
$$
and that $\Delta_i(t) \tendsto 0$ as $t \tendsto \infty$.
\end{mythm}
\begin{myproof}
Let 
$$
\ccA = \{\, i \suchthat \mbox{neuron-$i$ spikes infinitely often}\, \}
$$
($\ccA$ stands for ``active''), and
$$
\ccI = \{\, i \suchthat \mbox{neuron-$i$ stop spiking after a finite time} \,\}
$$
($\ccI$ stands for ``inactive'').
First consider $i \in \ccI$. Let $\spt{i}{k}$ be the time of the final spike. For any
$t > \spt{i}{k}$,
\begin{eqnarray*}
\avgcurr_i(t) - \currbias 
 & = & \frac{1}{t}\int_0^{\spt{i}{k}} (\curr_i(s)-\currbias)\,ds + 
       \frac{1}{t}\int_{\spt{i}{k}}^t (\curr_i(s)-\currbias)\,ds \\
 & = & \frac{1}{t}\int_0^{\spt{i}{k}} (\curr_i(s)-\currbias)\,ds + 
       \frac{1}{t}\poten_i(t) \\
 & = & \spikerate_i(t)\,\potenthres + \frac{1}{t}\poten_i(t), \\
\avgcurr_i(t) 
 & = & \spikerate_i(t)\,\potenthres + \currbias + \frac{1}{t}\poten_i(t).
\end{eqnarray*}
Note that $\poten_i(t) \le \potenthres$ always. If $\poten_i(t) \ge 0$, then
$$
0 \le \thres(\avgcurr_i(t))  - \spikerate_i(t) \le \potenthres/t.
$$
If $\poten_i(t) < 0$, 
$$
-\spikerate_i(t)\,\potenthres \le \thres(\avgcurr_i(t)) - \spikerate_i(t)\,\potenthres \le 0.
$$
Since $i\in\ccI$, $\spikerate_i(t) \tendsto 0$ obviously. Thus
$$
\thres(\avgcurr_i(t)) - \spikerate_i(t)\,\potenthres  \tendsto 0.
$$

\VS
Consider the case of $i\in\ccA$. 
For any $t > 0$, let $\spt{i}{k}$ be the largest spike time
that is no bigger than $t$. Because $i\in\ccA$, $\spt{i}{k} \tendsto \infty$ as $t\tendsto\infty$.
\begin{eqnarray*}
\avgcurr_i(t) - \currbias 
 & = & \frac{1}{t}\int_0^{\spt{i}{k}} (\curr_i(s)-\currbias)\,ds + 
       \frac{1}{t}\int_{\spt{i}{k}}^t (\curr_i(s)-\currbias)\,ds  \\
 & = & \spikerate_i(t)\,\potenthres + 
       \frac{1}{t}\int_{\spt{i}{k}}^t (\curr_i(s)-\currbias)\,ds.
\end{eqnarray*}
Furthermore, note that because of the assumption $\spt{i}{k+1}-\spt{i}{k} \le 1/r$ always,
where $r >0$, $\lim\inf \spikerate_i(t) \ge r$. In otherwords, there is a time $T$ large enough
such that $\spikerate_i(t) \ge r/2$ for all $i\in\ccA$ and $t\ge T$.
Moreover, $0 \le t - \spt{i}{k} \le \spt{i}{k+1} - \spt{i}{k} \le 1/r$ and 
$\curr_i(t) - \currbias \in [B_--\currbias, B_+-\currbias]$. Thus 
$$
\frac{1}{t}\int_{\spt{i}{k}}^t (\curr_i(s)-\currbias)\,ds 
\in \frac{1}{t} \, [B_--\currbias, B_+-\currbias] / r \tendsto 0.
$$
When this term is eventually smaller in magnitude than $\spikerate_i(t)\,\potenthres$,
$$
\thres(\avgcurr_i(t)
  =  \spikerate_i(t)\,\potenthres + 
       \frac{1}{t}\int_{\spt{i}{k}}^t (\curr_i(s)-\currbias)\,ds
$$
and we have
$$
\thres(\avgcurr_i(t)) - \spikerate_i(t)\,\potenthres  \tendsto 0.
$$
\qed
\end{myproof}

\section{Spiking Neural Nets and LCA}\label{sec:snn-lca}

This section shows that for a spiking neural net (SNN) that corresponds to a LCA, the
limit points of the SNN necessarily are the fixed points of the LCA. In particular,
when the LCA corresponds to a constrained LASSO, that is LASSO where the parameters are
constrained to be nonnegative, whose solution is unique, then SNN necessarily converges to
this solution. The proof for all these is surprisingly straightforward.

\VS
The following differential equation connecting $\dot{\avgcurr}_i(t)$ to $\avgcurr_i(t)$ and
all other spiking rates $\spikerate_j(t)$ is crucial.

\begin{equation}\label{eqn:DE-u-and-a}
\dot{\avgcurr}_i(t) = 
\frac{1}{\expwidth}\left( \inputcurr_i - \avgcurr_i(t) \right) 
 - \sum_{j\neq i} w_{i,j} \spikerate_j(t) 
 - \frac{1}{t}\left( \avgcurr_i(t) - \inputcurr_i \right). \tag{rates-DE}
\end{equation}

Derivation of this relationship is straightforward. First, apply the operation $(1/t)\int_0^t$ to
Equation~\ref{eqn:currents-DE}:
$$
\frac{1}{t}\int_0^t \dot{\curr}_i(s)\,ds =
\frac{1}{\expwidth}\left(\inputcurr_i - \avgcurr_i(t)\right) - \sum_{j\neq i} w_{i,j} \spikerate_j(t).
$$
To find an expression for the left hand side above, note that
\begin{eqnarray*}
\frac{d}{dt} \avgcurr_i(t) 
 & = & \frac{d}{dt}\; \frac{1}{t}\int_0^t \curr_i(s) \,ds  \\
 & = & \frac{1}{t} \curr_i(t) - \frac{1}{t^2} \int_0^t \curr_i(s)\,ds \\
 & = & \frac{1}{t} \left( \curr_i(t) - \avgcurr_i(t) \right).
\end{eqnarray*}
Therefore
\begin{eqnarray*}
\frac{1}{t} \int_0^t \dot{\curr}_i(s)\,ds
 & = & \frac{1}{t}\left( \curr_i(t) - \inputcurr_i \right) \\
 & = & \frac{1}{t}\left( \curr_i(t) - \avgcurr_i(t) \right) +
       \frac{1}{t}\left( \avgcurr_i(t) - \inputcurr_i \right) \\
 & = & \frac{d}{dt} \avgcurr_i(t) + \frac{1}{t} \left( \avgcurr_i(t) - \inputcurr_i \right).
\end{eqnarray*}
Consequently, Equation~\ref{eqn:DE-u-and-a} is established.

\VS
Observe that because $\curr_i(t)$ is bounded (Proposition~\ref{prop:curr-spikerate-bounds}), so
is the average current $\avgcurr_i(t)$. This means that $\dot{\avgcurr}_i(t) \tendsto 0$ as
$t \tendsto \infty$ because it was shown just previously that
$\dot{\avgcurr}_i(t) = (\curr_i(t)-\avgcurr_i(t))/t$. 

\VS
Since $\curr_i(t)$ and $\spikerate_i(t)$ are all bounded, 
the vectors $\vvu(t)$ must have a limit point (Bolzano-Weierstrass) $\vvustar$.
By Theorem~\ref{thm:a-and-u}, there is a correpsonding $\vvastar$ such that
$\vvthres(\vvustar) = \vvastar\potenthres$.
Moreover, we must have
$$
\vvzero = \frac{1}{\expwidth} (\vvb - \vvustar) - W\vvastar
$$
where the matrix $W$ has entries $w_{i,j}$ and $w_{i,i} = 0$. 
Hence
$$
\vvzero = \frac{1}{\expwidth}(\vvb - \vvustar) - \frac{1}{\potenthres} W \thres(\vvustar).
$$
Indeed, $\vvustar$, $\vvastar = \thres(\vvustar)$ correspond to a fixed point of LCA.
In the case when this LCA corresponds to a LASSO with unique solution, there is only one fixed point,
which implies that there is also one possible limit point of SNN, that is, the SNN must converge, and
to the LASSO solution.

\small
\bibliographystyle{abbrv}
\bibliography{snn}

\end{document}